# Explanation of Unintended Radiated Emission Classification via LIME

Tom F. Grimes[1], Eric D. Church[2], William K. Pitts[3], and Lynn S. Wood[4]

*Abstract--* Unintended radiated emissions arise during the use of electronic devices. Identifying and mitigating the effects of these emissions is a key element of modern power engineering and associated control systems. Signal processing of the electrical system can identify the sources of these emissions. A dataset known as Flaming Moes includes captured unintended radiated emissions from consumer electronics. This dataset was analyzed to construct next-generation methods for device identification. To this end, a neural network based on applying the ResNet-18 image classification architecture to the short time Fourier transforms of short segments of voltage signatures was constructed. Using this classifier, the 18 device classes and background class were identified with close to 100 percent accuracy. By applying LIME to this classifier and aggregating the results over many classifications for the same device, it was possible to determine the frequency bands used by the classifier to make decisions. Using ensembles of classifiers trained on very similar datasets from the same parent data distribution, it was possible to recover robust sets of features of device output useful for identification. The additional understanding provided by the application of LIME enhances the trainability, trustability, and transferability of URE analysis networks.

*Index Terms--* URE, STFT, CNN, explainability, LIME, interpretability

## I. Introduction

IN this paper, unintended radiated emissions (UREs) will be loosely defined as electromagnetic emissions radiated by devices during their operation. UREs result from a mismatch between ideal and real designs. Due to this mismatch, UREs form a lost energy pathway along with heat, vibration, and other losses. URE energy includes emitted radio frequency (RF) as well as RF that is conducted over the power infrastructure. As a result, UREs form a very useful signal for measuring activity without being immediately next to the emitting device.

Broad categories for analysis based on the characterization of URE signatures include electromagnetic interference (EMI), nonintrusive load monitoring (NILM), and information security (IS) [1]. Loosely, EMI research is focused on reducing the magnitude of the URE signature for the fabrication of better devices, NILM research is focused on characterizing the signature for load monitoring and maintenance applications, and IS research is focused on exploiting the URE by obtaining information that it reveals about the device operation on the circuit [1].

NILM for device identification is finding application in the consumer market [2]. NILM is advertised as a means of reducing power bills by identifying what devices are running and identifying those that represent an outsized contribution to the household energy use. NILM measurements can consist of current data, voltage data, or a combination of both.

## II. Data

In 2016, Oak Ridge National Laboratory generated the Flaming Moes dataset [1] as an idealized URE dataset that can be used for the development of URE detection algorithms. The dataset (as provided) includes two non-consecutive 10-minute segments of voltage data captures measured using a high impedance differential voltage probe between the ground and neutral conductors with each of 18 devices (Table I) in an "on" state and four 10-minute data captures with the device in an "off" state. The fluorescent light "off" state was selected as the background state for classification because it was the only device to be plugged in but turned off for all device "off" states. Devices were given a minimum of 1 minute to stabilize between being plugged in and the beginning of the capture. The URE signal was recorded using an Ettus Research LFRX analog-to-digital processing board with one channel at 2 MS/s. The collection system also featured a twin T-notch filter used to remove the 60 Hz component of the signal taken from the ground and neutral lines of the plug.

TABLE I: Device List

| | |
|---|---|
| Background | Polycom VoIP |
| Corelco Phone | Raspberry Pi |
| CyberPower UPS | Roku 2 XS |
| Dell Monitor | USRP E310 |
| Dell Optiplex | ViewSonic Monitor |
| Dell XPS | Vizio Blu-ray |
| Fluorescent Lights | VTech V.Smile |
| LG Phone | Wii U |
| Linksys Router | Xbox One |
| Odroid XU4 | |

## III. Network

For classification of the URE signatures of the various devices, the data were first transformed by a short time Fourier transform (STFT). STFT transforms for the analysis and neural classification of 1D signals are popular techniques with successful applications in areas including but not limited to: URE detection [3], electrocardiogram (EKG) analysis [4], nuclear detector particle identification [5], and audio identification [6].

[1] Pacific Northwest National Laboratory, P.O. Box 999, Richland, Washington 99352 (e-mail: thomas.grimes@pnnl.gov)
[2] Pacific Northwest National Laboratory, P.O. Box 999, Richland, Washington 99352 (e-mail: eric.church@pnnl.gov)
[3] Pacific Northwest National Laboratory, P.O. Box 999, Richland, Washington 99352 (e-mail: Karl.Pitts@pnnl.gov)
[4] Pacific Northwest National Laboratory, P.O. Box 999, Richland, Washington 99352 (e-mail: Lynn.Wood@pnnl.gov)

STFT analysis offers some concrete benefits over alternatives: STFT pre-processing allows the network to converge very rapidly (especially when compared to time domain-only solutions, which are not given frequency domain information and must therefore learn it). STFTs allow data compression for data taken at high rates with minimal loss of useful information. In this instance, $2 \times 10^5$ points are transformed to $5 \times 10^4$. STFTs allow the data to be presented in a much more human-interpretable manner than time-series data. Finally, STFTs retain time domain information (unlike fast Fourier transform-only representations).

To classify the STFTs, a standard image classifier network, ResNet18, was used [7]. This network was chosen as a tested and accepted network architecture for image classification. Because the STFT generates 2D data, it is possible for the STFT algorithm to directly generate the shape of the input required for the image classification network. The image size chosen was 224 x 224, which was the default input shape for ResNet18. Parameters of the STFT algorithm were chosen such that the output matched this shape.

---

**Algorithm 1:** Training Data Generation Algorithm

**Input:**
$W_{FFT}$ – # of points to be used in the FFT
$W_D$    – # of points in the dataset
$f_{Tr}$  – percent of data used for training
$C$      – set of data classes

**Output:**
$I$  – array of spectrograms to generate

**for each** i in I:
$c \leftarrow random \in C$
$s \leftarrow random \in [0, (W_D(c) - 2W_{FFT}) * f_{Tr} - 1]$
$i \leftarrow stft(data_c(s, s + W_{FFT}))$

---

Training data were generated using the algorithm described in Alg. 1. Spectrograms were generated by randomly selecting a class and a starting point in the data. Training data starting points were taken from the interval $[0, (W_D(c) - 2W_{FFT}) * f_{Tr} - 1]$. Validation data were generated by selecting start points in $[(W_D(c) - 2W_{fft}) * f_{Tr} - 1 + W_{fft}, W_D(c) - 1 - W_{fft}]$. Spectrogram width was set to 0.1 s of data ($2 \times 10^5$ samples at 2 MHz). The spectrograms were generated by the STFT methods in Librosa [8] or SciPy [9], with the outputs shown to be equivalent. Spectrogram parameters were chosen such that the output dimensions would match the expected input dimensions for the image classifier of 224 x 224. This plane was duplicated to form the 224 x 224 x 3 input expected by ResNet18. Training was found to perform equally well both on the raw output of the STFT methods and on the log of the output. Because the log output is more human-visible and -understandable, those are the networks for which results will be displayed in this paper.

Across repeated trainings, the network starting with randomly initialized weights rapidly converged to accuracies generally exceeding 95 percent and often as high as 100 percent after 10 epochs of training. A standard training accuracy curve and confusion matrix appear in Figs. 1 and 2.

Looking at the sample STFT images in Fig. 3, it is not surprising that the algorithm is able to separate them with nearly 100 percent accuracy after a small amount of training. The class representative spectrograms look unique and identifiable by eye (with the possible exception of a few classes that strongly resemble background).

Despite this strong classification performance, an end-user classifying emissions with this network as a black box would be missing critical information that would be strongly desired by a scientist studying the URE. There is no description of what frequencies are being used by the network. There is no indication of what other conditions this network will be valid for or what backgrounds it will be robust against. For this information we must turn to an interpretability code: our choice for this task is LIME.

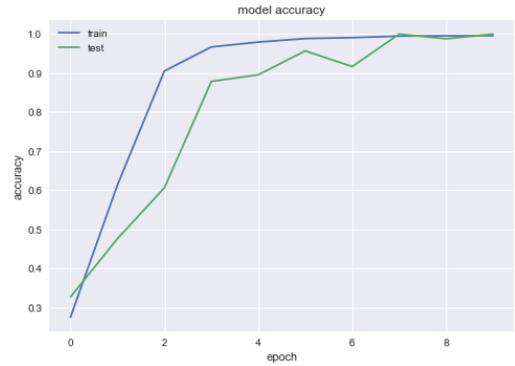

Fig. 1. Representative train and test accuracy curves for training of the STFT/CNN network on Flaming Moes.

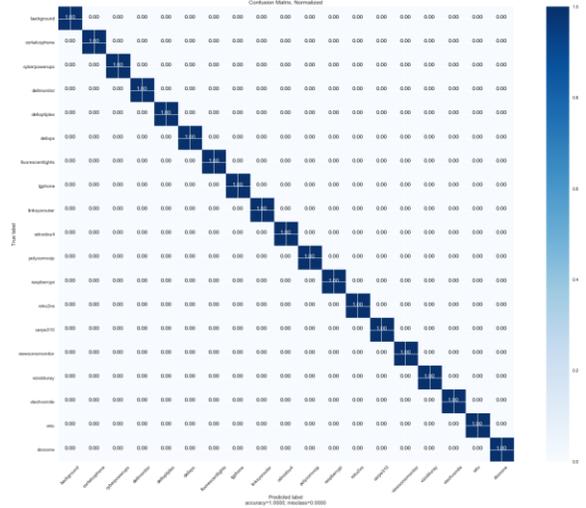

Fig. 2. Representative confusion matrix from a trained STFT/CNN on the Flaming Moes data.

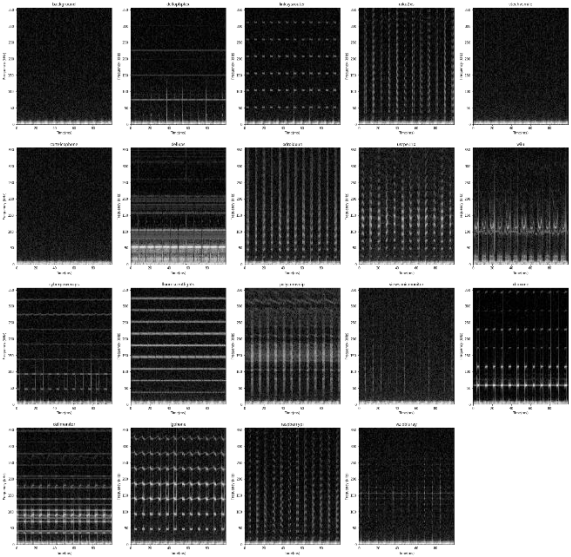

Fig. 3. STFTs generated from each of the 19 Flaming Moes classes.

## IV. LIME [2]

The Local Interpretable Model-agnostic Explainability (LIME) code was written by researchers at the University of Washington. LIME is available on Github [10].

LIME works by building a local linear approximation of a complex model's behavior in the neighborhood of an example by treating the model as a black box and classifying near permutations of the example being explained. Using that local model, the inputs most important to classification are identified. Because the model is treated as a black box, LIME's explanations are not limited to explaining a single class of model and can instead be applied to any model that makes predictions on the input.

A vivid example of the usefulness of LIME is described in [10]. A binary classifier was constructed using a network to distinguish between two classes (wolf and dog). While the classifier performed extremely well on the training data, it performed poorly when used in practice. Through the superpixels it identified as important, LIME was able to show that the network classification was based on pixels belonging to the background rather than the signal.

This level of insight is incredibly important. It is easy to imagine a situation where there is a positive correlation between a class of data and a characteristic background. LIME can help to design and verify networks are using components of signal rather than noise to make decisions. For instance, if data from a URE-emitting device can only be obtained in a single location with a fixed background, then combining that data with data taken with a different background poses substantial issues that LIME can help rectify.

LIME uses an image processing algorithm known as quickshift [11] to separate the image into "superpixel" segments. The algorithm uses these superpixels to create near permutations of the example image. Quickshift works by creating a tree of nodes dictated by the distance between nearby nodes in both pixel and color values. Quickshift has three hyperparameters. First, kernel size sets the size of the neighborhoods considered for the pixels. Second, the maximum distance sets a maximum cutoff in distance (this is done mostly for computational efficiency and is generally set to infinity or as small multiple of the kernel size). Third, multiplying the spectrogram pixel values by a constant changes the relative importance of distance matching and color matching in the quickshift algorithm. Large multipliers result in very fragmented regions strongly adhering to matching color (as in Fig. 4) whereas small multipliers result in large blocky regions that are generally constrained only by strong color boundaries. A standard 224 x 224 image is decomposed by the algorithm into on the order of 200 superpixels.

For each image being explained, LIME is able to either identify the top N superpixels, which support the case that the image belongs to a given class, the top N superpixels supporting the case that the image does not belong to a given class, or the most influential N superpixels for classification in general. For this work, LIME recorded the top three superpixels supporting the decision that the spectrogram being explained belongs to the class corresponding to that spectrogram's ground truth (even in the cases where the model classification was incorrect).

Fig. 4 shows the superpixels LIME identified to support that the spectrogram in the figure with ground truth Wii U belongs to the Wii U class. This individual explanation is reasonably enlightening. LIME has identified the u-shaped catenary curves around 100 kHz with the region on the left. LIME has identified a region touching both the low frequency band and the catenary band with the region on the right. Finally, the region at the top identifies the characteristic noise or lack of information at high frequency. Despite this relatively interpretable spectrogram, it would be difficult for a human to decide what things were important—especially if the spectrogram had additional extraneous background features. To increase the interpretability, aggregation of explanation superpixels was performed on a large number of spectrograms of the same class.

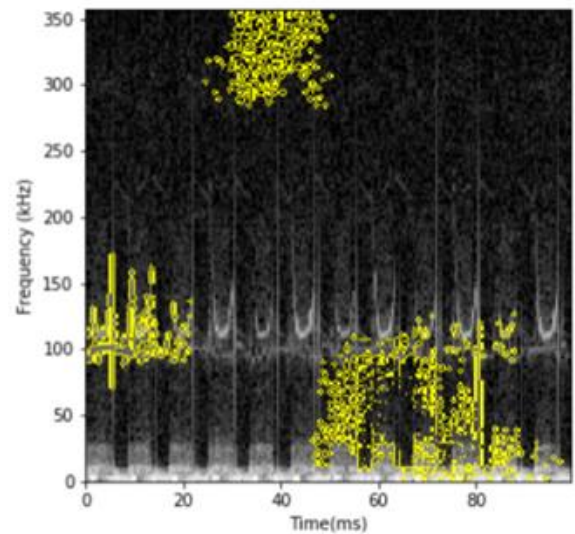

Fig. 4. LIME explanation for classification of STFT as Wii U.

## V. AGGREGATIONS

The motivation behind the aggregation is that, while a single image provides information that may or may not be human-interpretable, the superposition of many such explanations may be more interpretable by humans. Unlike the common neural net application of image classification, the interesting features of the FFT images are locked to a single horizontal frequency band. Thus, aggregation of the explanation process causes identified features to stack up at interesting frequencies. Because the starting location is not locked in phase to the 60 Hz signal, the characteristic features of the device can occur anywhere along the timeline. As such, it would be expected that identified regions would form horizontal bars of varying intensity once the explanations are summed over a large number of spectrograms. The process of generating the aggregation is described in Alg. 2.

The expected horizontal banding is exactly what is seen by the aggregated explanation for spectrograms with ground truth Wii U shown in Fig. 5. This figure was generated by finding the identified regions from each of 400 spectrograms (i.e., the yellow bounded regions from Fig. 4). Then the aggregation was formed by summing the number of times each pixel occurred in identified regions. Finally, the values were normalized to fall in [0,1].

**Algorithm 2:** LIME Output Aggregation
**Input:**
  $M$   – explainer for model M
  $S$   – array of spectrograms to be explained
**Output:**
  $E$   – aggregated model explanation
**for each s** in S:
  $e \leftarrow M(s)$
  **for each pixel p** in e:
    E(p) += e(p)

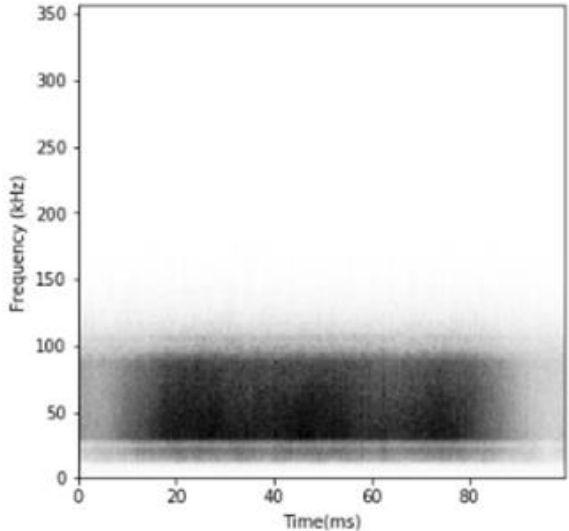

Fig. 5.  Aggregation of LIME explanations for 400 Wii U STFTs.

Because the horizontal time component of the explanation map does not carry any meaningful information, the time axis can be integrated over to form a representation that exists only in frequency space. Fig. 6 shows a comparison of the integrated LIME map for Wii U with the resultant of the STFT integrated over the time axis. Such a result is similar to a FFT but is actually the average of FFTs taken over subsets of the data. This transform is known as Welch's method [12]. The data transformed in this manner have broader peaks and generally smoother behavior leading to a very similar signal-to-noise ratio to the direct FFT.

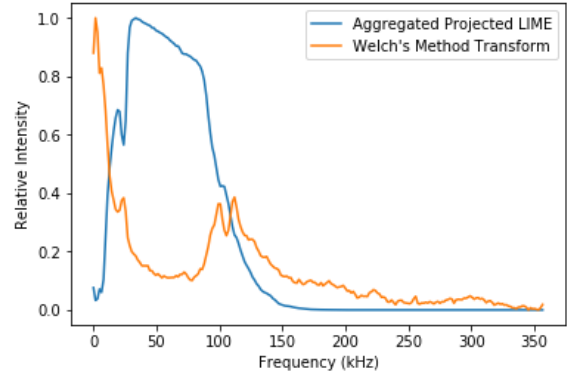

Fig. 6.  Welch's Method Transform and Aggregated LIME Frequency Projection for Wii U.

As is apparent, there were a large number of Wii U spectrograms with identified superpixels in the region between the ~25 kHz features and the ~100 kHz features, much like the region in the bottom right of Fig. 4. This is evident from the thick banding in Fig. 5 that touches on both of those bands as well as from the large intensity in the Aggregated Projected LIME in Fig. 6 in between those bands (the bands themselves are evident in the Welch's Method Transform in Fig. 6).

From these facts, it becomes clear that this region is being selected by LIME not because the center of the region carries information, but rather because the frequency bands at the top and bottom of the identified region do. This is a result of the quickshift segmentation algorithm that generates the superpixels used by LIME. Superpixels are grouped by like Parzen density which is a function of pixel location and color value [11]. Thus, large changes in color values are preferentially found at or just beyond the edges of superpixels. Knowing this information, it stands to reason that the actual important frequencies would be identified by a change in intensity in the LIME aggregation rather than the magnitude of the intensity.

Fig. 7 shows the comparison between the Welch's method plot for the Wii U and the absolute value of the derivative of the projection of the aggregated LIME explanation onto the frequency axis. This explanatory distillation seems to indicate very strongly the feature usage by the network. The features that are identified show a very strong connection to real spectral features (compare the spikes in Derivative Aggregated Projected LIME to spikes in Welch's Method frequencies for the device). The process for generating the Derivative Aggregated Projected LIME from the Aggregated LIME is detailed in Algorithm 3.

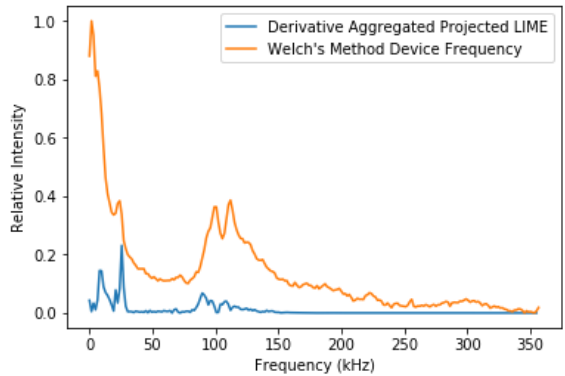

Fig. 7. Welch's Method Transform and Derivative Aggregated LIME Frequency Projection for Wii U.

---

**Algorithm 3:** Derivative Welch's Method
**Input:**
$E$ – Aggregated Model Explanation
**Output:**
$W_D$ – Explanation Welch's Method Derivative
**for each row** in E:
$W \leftarrow \sum E(row)$
**for each w** in W-1:
$W_d(w) = |W(w+1) - W(w)|$

---

The same process that was applied to the Wii U in Fig. 7 was applied to all the device classes in Fig. 8. Across the results for these devices, several interesting observations can be made. Devices can be identified by the network using a subset of their spectral features rather than all of them. Just as the class lines between a blue circle and a red square can be drawn many different ways, so too can the representations of these classes be divided by a subset of features. This LIME method generally identifies the outer edge of the band rather than the center—again this is a consequence of the superpixeling algorithm. As a result, the frequency resolution becomes a function of both the pixel size in the frequency dimension and the superpixeling parameters.

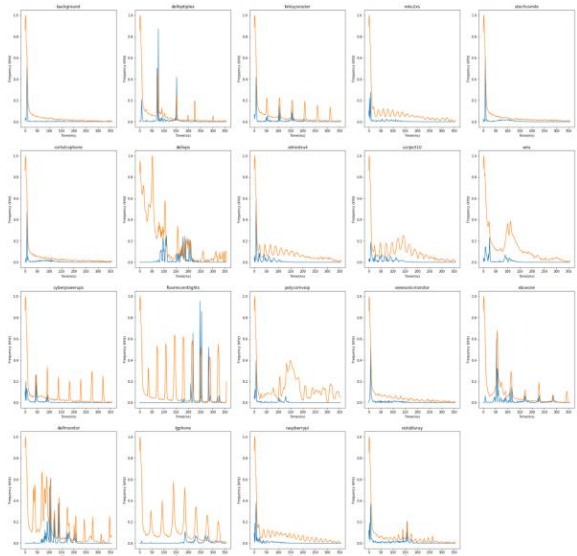

Fig. 8. Welch's Method Transform (orange) and Derivative Aggregated LIME Frequency Projection for all device classes (blue).

## VI. ENSEMBLES

One of the major takeaways from the graphs in Fig. 8 is that the network does not need to use a full set of the device features to make its classification and therefore it does not. It would be desirable if it were possible to recover the full set of features of the device to facilitate positive identification of the device. This would enhance network trustability by helping to ensure the network has identified the desired device.

It is possible to run the data generation process in Alg. 1 with a different random seed and achieve a different set of spectrograms built from the same parent distribution. Networks (even networks with identical architecture) built from the new set of spectrograms would be expected to have different weights after training. In fact, even networks with identical architecture trained on an identical dataset would be expected to have different weights if the data were presented to the training algorithm in a different order (resulting in the batches used for updating having a different makeup and a different state of the gradient to update against).

Applying LIME to large numbers of networks trained on spectrograms generated from the same parent distribution reveals that the aggregation of spectrogram explanations for the same piece of equipment (Wii U) varies significantly despite the fact that the overall network classification accuracy is extremely high (greater than 95 percent) for almost all of the networks. Given the previous revelation that the networks do not make use of the full set of device features when identifying classes, it is not surprising that different trainings of the network end up with different subsets of the features being used. Visually, the networks in Fig. 9 fall into a few major groups by the way that they identify the Wii U class. This is likely an indicator of the features being used to detect the class.

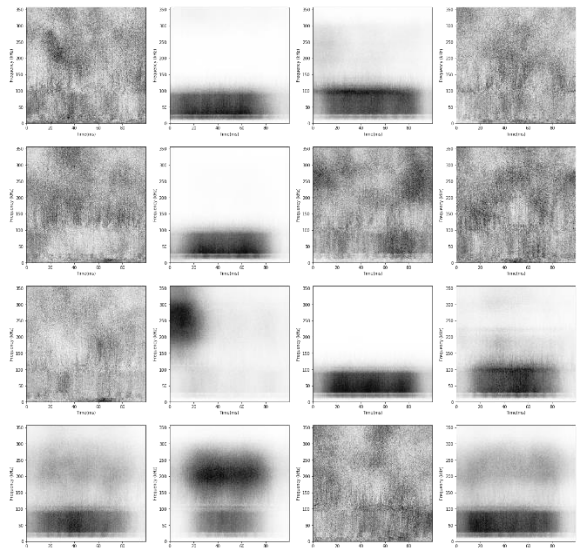

Fig. 9. Aggregated LIME explanations for Wii U across 16 retrainings of the same network architecture.

Another interesting phenomenon discovered while looking at many explanations across networks is the emergence of classes that "side train" and show significant structure in the time dimension (column 2, rows 3 and 4 of Fig. 9 are both exemplars). Strong temporality is somewhat rare (less than 10 percent of cases) and not noticeably deleterious to network accuracy over this dataset, but could reasonably inhibit transferability. Even among networks that exhibit this phenomenon, it is common that only one or two of the classes will exhibit strong temporality and the rest will be temporally homogenous. Two predominant theories explain this behavior. One theory is that the network has trained on a feature with part of the overall pattern obscured and therefore is only able to find it at the edge of the image. The other theory maintains that the network has learned a long-range (in time) association of features, but LIME is only expressing the behavior at the location of one of the sub-parts of the pattern. As a result, it always finds that sub-part in the same location because the other sub-part's relative location constrains the location of the part being found.

Using Algorithm 3, the plots from Fig. 9 can be transformed from the 2D aggregated LIME maps to 1D descriptions of frequencies used for identification by the network. These plots are shown in Fig. 10. The Welch's method transform of the device signal is plotted alongside for comparison. The various plots show that the frequency band usage between the networks is significantly more similar than the 2D LIME maps. Seemingly different formations of mass in the 2D plane are revealed to be indicative of very similar feature sets once the absolute value of the projection's derivative is taken. In fact, there appears to be large agreement on the frequency bands of importance among the networks.

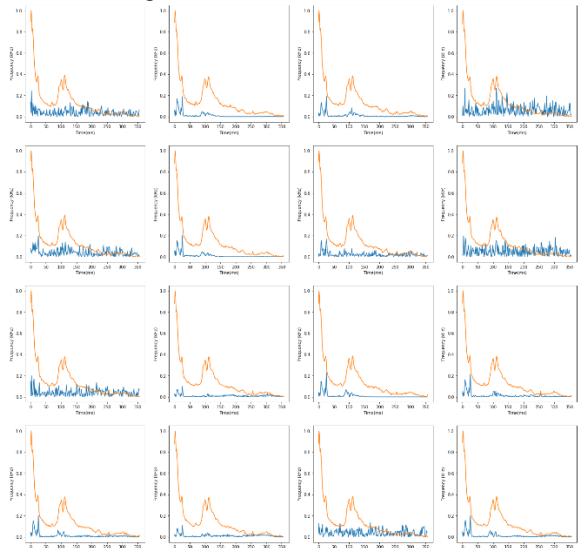

Fig. 10. Welch's Method Transform (orange) and Derivative Aggregated LIME Frequency Projection for Wii U for 16 re-trainings of the same network architecture (blue).

However, a large difference exists in the noise background. Plots with large amounts of intensity at frequencies not corresponding to features in the FFT all come from the same family of 2D plots where the mass in 2D space is very distributed. While the peaks of interest are still generally identifiable, the background that they appear against is significantly greater. The accuracy of the overall network does not appear to be correlated with the family that the LIME map for the Wii U class falls into.

The authors theorize that networks using a noisy set of features are still able to train to very high accuracy because the classes in the dataset are easily distinguished by the network and therefore many different subsets of features can be used to make the same classification distinction. Additionally, a component of noise in the feature set is induced not by the inner workings of the network but rather as an artifact of the first order numerical differentiation instantiated in Alg. 3 as well as noise generated when the edges of superpixels do not border on frequencies of interest.

Fig. 11 shows the mean and standard deviation for the plots in Fig. 10. This distribution was created by combining explanations from many networks built from the same parent data distribution and arrives at a representation that simultaneously suppresses the background noise that appears in some of the plots and identifies a larger set of real features than many of the individual plots did. It is not unreasonable to believe that this process of aggregating solutions from networks trained on the same data might approach at the limit the identification of the full feature set of features useful for separating this device from other device classes. In a noisy environment, this process may still be able to recover components of the device signal that are extremely difficult to identify in the background containing FFT. The feature set generated by this process is also an ideal feature set to be used to compare to the FFT and known data about device behavior to ensure that the network is using true features in the device identification rather than background.

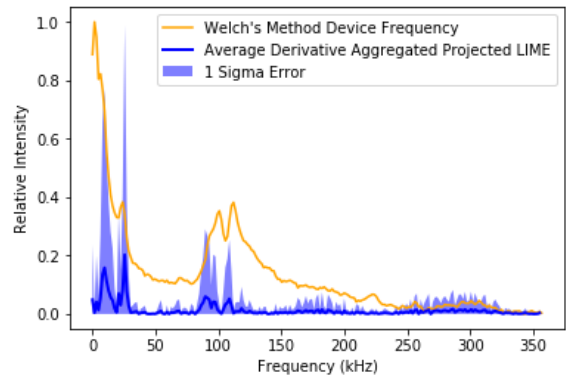

Fig. 11. Welch's Method Transform and Average Derivative Aggregated LIME Frequency Projection for Wii U.

## VII. CONCLUSION

This work has shown a method to use URE signatures for nonintrusive load monitoring applications. The STFT/ResNet18 network design was extremely efficient for generating classifiers that were able to discriminate between the device classes with a high degree of accuracy.

LIME was used to successfully identify sections of images that were important in the classification of the STFT images. Through transformation and aggregation performed on the LIME explanations for individual images it was shown that the regions identified by LIME correspond to true features in the output of the device. Furthermore, it was shown that these features are identified in a human-interpretable fashion. It was also shown that there is potential usefulness in these frequency explanations for device identification. This frequency characterization improved in quality with the ensembling of additional models of the same architecture.

The additional network explainability afforded by this work makes formerly black box models for classifying UREs much more understandable. With this understanding the models become more trainable, trustable, and transferable because it is possible to inspect the features used by the model and ensure that they are capturing meaningful information, correspond to features of the equipment, and are not strongly correlated with irrelevant background signatures.

## VIII. FUTURE WORK

The first most obvious additional task to be performed is to test these techniques on less clean, more realistic real-world data. A rigorous investigation containing key insights and network development can be found in [13].

Next, it would be very interesting to begin tuning the internals of LIME to be more suited to this specific problem. That tuning might include modifying the segmentation algorithm to enforce segmentation more conducive to the generation of horizontal frequency bands. Tuning might also involve modifying the image explanation subclass to work more efficiently or on numbers of planes other than 1 and 3 to allow for the application of additional network techniques. Further investigations on the side training networks to identify the root cause of the phenomenon would also be interesting.

Finally, it is extremely interesting to continue investigating model ensembling to determine to what extent the features of a device can be recovered. In a very noisy environment where the device features could not be easily determined from the FFT, it would be extraordinarily useful if the features identified by the ensemble could reproduce the device signal output.

### ACKNOWLEGEMENT

This work was funded by the U.S. Department of Energy National Nuclear Security Administration's Office of Defense Nuclear Nonproliferation Research and Development (NA-22).

We wish to acknowledge our Pacific Northwest National Laboratory colleagues for their important discussions: Aaron Tuor, Alex Hagen, Aaron Luttman, Eva Brayfindley, Luke Erikson, and Mark Greaves.

### REFERENCES

[1] A. M. Vann, T. P. Karnowski, R. Kerekes, C. D. Cooke, and A. L. Anderson, "A Dimensionally Aligned Signal Projection for Classification of Unintended Radiated Emissions," *IEEE Trans. Electromagn. Compat.*, vol. 60, no. 1, pp. 122-131, Apr. 2017.
[2] Sense, Accessed February 5, 2020, Available: https://sense.com/product.
[3] Jason M. Vann, Adam L. Anderson, and Thomas Karnowski, "Classification of Unintended Radiated Emissions in a Multi-Device Environment," *IEEE Transactions on Smart Grid*, vol. 10, no. 5, Sept. 2019.
[4] Sara S. Abdeldayem and Thirimachos Bourlai, "ECG-based Human Authentication using High-level Spectro-temporal Signal Features," Presented at 2018 IEEE International Conference on Big Data, Seattle, Washington. DOI: 10.1109/BigData.2018.8622619.
[5] Brian Archambault, Tom Grimes "Discriminating nuclear recoils from alpha particles using acoustic signatures in γ / β -blind tensioned metastable fluid detectors" poster presentation - SORMA XVII – June 11, 2018
[6] Awni Hannun, Carl Case, Jared Casper, Bryan Catanzaro, Greg Diamos, Erich Elsen, Ryan Prenger et al. "Deep speech: Scaling up end-to-end speech recognition." Available: arXiv:1412.5567
[7] Kaiming He, Xiangyu Zhang, Shaoqing Ren, and Jian Sun, "Deep Residual Learning for Image Recognition," Available: arXiv:1512.03385 [cs.CV].
[8] LibROSA, Accessed February 5, 2020 at https://librosa.github.io/librosa/.
[9] Scipy Accessed February 5, 2020 at https://www.scipy.org/.
[10] Ribeiro, Marco Tulio, Sameer Singh, and Carlos Guestrin. "Why should I trust you?: Explaining the predictions of any classifier." In Proceedings of the 22nd ACM SIGKDD international conference on knowledge discovery and data mining, pp. 1135-1144. ACM, 2016.
[11] Brian Fulkerson and Andrea Vedaldi, Quick shift image segmentation, Accessed February 5, 2020 at https://www.vlfeat.org/api/quickshift.html.
[12] P. D. Welch, "The use of fast Fourier transforms for the estimation of power spectra: A method based on time averaging over short modified periodograms," *IEEE Trans. Audio Electroacoust.*, vol. 15, pp. 70-73, 1967.
[13] Tom Grimes, Eric Church, Karl Pitts, Lynn Wood "Adversarial Training for URE Classification"

## IX. BIOGRAPHIES

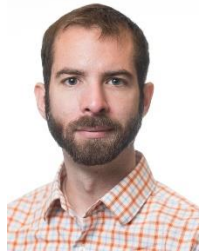

**Tom Grimes** (M'12) received his PhD in Nuclear Engineering and Masters in Business Administration from Purdue University in 2015.

He is a Physicist at PNNL. Prior to working at PNNL, he was an IC Postdoctoral fellow and a NSF Graduate Fellow at Purdue University.

Dr. Grimes' research interest focus on deep learning with an emphasis on explainability. Applications of interest include electromagnetics, particle spectroscopy, and metastable fluids.

**Eric Church** received his PhD from the University of Washington, Seattle, WA, in Experimental Particle Physics in 1996.

He is a Physicist at PNNL. Prior to his arrival at PNNL, he was a Research Associate Scientist with Yale University and a post-doctoral researcher with UC-Riverside.

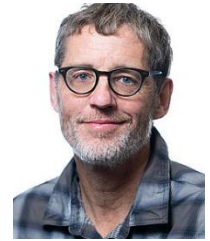

Dr. Church leads various internal lab investments and DOE projects including research into neutrino oscillations and the majorana nature of neutrinos. Current experimental neutrino collaborations on which he works include MicroBooNE, DUNE and NEXT.

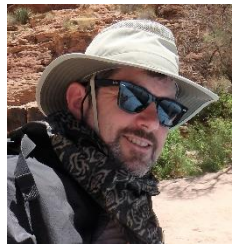

**William Karl Pitts** received his PhD in Nuclear Physics from Indiana University in 1987. He is a Physicist at PNNL. Prior to joining PNNL in 2000 he was an Associate Professor from the University of Louisville and a Research Scientist at the University of Wisconsin. His current interests include signal processing, EM/RF studies, and material processing.

**Lynn Wood** received his PhD from the University of California, Davis in Nuclear Physics in 1998, after which he was a post-doctoral researcher at Iowa State University before spending 11 years in the embedded systems industry, where he managed ASIC processor design development. He is currently a Physicist at PNNL.

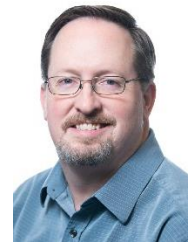

Dr. Wood leads PNNL's involvement in the Belle II high-energy physics experiment based in Japan, as well as several projects sponsored by National Nuclear Security Administration.